\documentclass[letterpaper, 10 pt, conference]{ieeeconf}  

\IEEEoverridecommandlockouts                              

\usepackage[pdftex]{graphicx}
\graphicspath{{figures/}}
\DeclareGraphicsExtensions{.pdf,.jpeg,.png}

\usepackage{amsmath} 
\usepackage{amssymb}  

\usepackage{cite}
\usepackage{array}

\usepackage{multirow}
\usepackage{hhline}
\usepackage{makecell}
\usepackage[font=footnotesize]{caption}
\usepackage{subcaption}
\usepackage[T1]{fontenc}
\usepackage[utf8]{inputenc}
\usepackage{xcolor}

\newcolumntype{?}{!{\vrule width 1pt}}
\newcommand{\size}[2]{{\fontsize{#1}{0}\selectfont#2}}

\begin{document}

\title{\LARGE \bf
Integrating Deep Reinforcement and Supervised Learning to Expedite Indoor Mapping
}

\author{Elchanan Zwecher$^{1}$, Eran Iceland$^{2}$, Sean R. Levy$^{3}$, Shmuel Y. Hayoun$^{4}$, Oren Gal$^{5}$ and Ariel Barel$^{6}$
\thanks{$^{1}$Elchanan Zwecher is with the Computer Science Department,
        Hebrew University of Jerusalem, Jerusalem, Israel
        {\tt\small elchanan4567@gmail.com}}%
\thanks{$^{2}$Eran Iceland is with the Computer Science Department,
        Hebrew University of Jerusalem, Jerusalem, Israel
        {\tt\small eran.iceland@gmail.com}}%
\thanks{$^{3}$Sean R. Levy is with the Faculty of Aerospace Engineering, Technion - Israel Institute of Technology, Haifa, Israel
        {\tt\small sean2102@gmail.com}}%
\thanks{$^{4}$Shmuel Y. Hayoun is with the Faculty of Aerospace Engineering, Technion - Israel Institute of Technology, Haifa, Israel
        {\tt\small shmuli.hayoun@gmail.com}}%
\thanks{$^{5}$Oren Gal is with the Geo-information Department,
        Technion - Israel Institute of Technology, Haifa, Israel
        {\tt\small orengal@alumni.technion.ac.il}}%
\thanks{$^{6}$Ariel Barel is an academic visitor at the Computer Science Department, Technion - Israel Institute of Technology, Haifa, Israel {\tt\small arielba@technion.ac.il}}%
}

\maketitle
\thispagestyle{empty}
\pagestyle{empty}

\begin{abstract}

The challenge of mapping indoor environments is addressed. Typical heuristic algorithms for solving the motion planning problem are frontier-based methods, that are especially effective when the environment is completely unknown. However, in cases where prior statistical data on the environment's architectonic features is available, such algorithms can be far from optimal. Furthermore, their calculation time may increase substantially as more areas are exposed.
In this paper we propose two means by which to overcome these shortcomings. One is the use of deep reinforcement learning to train the motion planner. The second is the inclusion of a pre-trained generative deep neural network, acting as a map predictor. Each one helps to improve the decision making through use of the learned structural statistics of the environment, and both, being realized as neural networks, ensure a constant calculation time. We show that combining the two methods can shorten the duration of the mapping process by up to 4 times, compared to frontier-based motion planning.

\end{abstract}

\section{INTRODUCTION}
Indoor mapping by an autonomous agent is a practical task that has been researched for many years. We are given an agent with navigation peripheral range-sensing capabilities. The agent aims to construct a map of its environment through exploration in minimal time. In terms of the motion planning, the goal is to find an optimal strategy that will allow the agent to complete this task. Indoor mapping can be categorized into one of two types, corresponding to the available information about the environment:\\
\textbf{Mapping an unknown environment:} There is no information about the structure, except perhaps its boundary, and the fact that all rooms are accessible. This problem is in fact the easiest problem to optimize, since there is little knowledge to rely on. It can be solved using frontier-based algorithms, by which the agent always moves towards a chosen point on the boundary ("front") between the observed and unobserved areas. Each of these algorithms is distinguished by its point selection logic (e.g. the nearest point, the point with the highest exposure potential, etc.). These front-based algorithm are guaranteed to complete the mapping process.\\
\textbf{Mapping a partially-known environment:} Although the specific environment is unknown, the statistics of its architecture are available to the exploring agent. In this case, this knowledge may utilized to expedite the mapping process. This case is probably more common than one might expect, since most buildings share typical properties, such as straight continuous and perpendicular walls, continuous spaces, relatively rectangle rooms, wide enough corridors, etc. Here heuristic frontier-based algorithms that do not incorporate any knowledge of the structural statistics may be far less effective than in the previous case.

Our present work focuses on the second type of problem. In order to integrate the statistics into the motion planning it is possible to manually derive a set of intuitive rules (straight walls, corners, etc.). However, this is limited to a human's capability of interpreting large and fairly abstract amounts of data into concise correlations between different architectonic features. In the present work we propose the use of deep learning (DL) and deep reinforcement learning (DRL) techniques in order to incorporate any available statistical data more effectively.



Machine learning (ML) in general involves the process of automatically extracting insights from statistical data for the purpose of decision making. The field first addressed tasks such as image classification, voice recognition and others which involved providing one-time solutions. In recent years the field has also broken out in the direction of RL for solving serial problems, that is of setting a strategy or policy in which each decision influences the next decision \cite{mnih2015humanlevel} \cite{sutton2018reinforcement}. RL has been extensively studied as a policy generator for games such as Atari, Chess, Go and Backgammon. In these tasks, the agent's early decision affects its subsequent decisions and, ultimately, the final result. Learning by reinforcement in gaming has shown impressive results \cite{silver2016mastering} \cite{silver2017mastering}, better than algorithms which are not based on an offline study of the statistical properties of data available a priori, but rather on running online simulations of the specific task (e.g. rule-based methods or online statistical-based methods, like Monte-Carlo).
The success of RL in gaming is due to the fact that extensive training can be performed offline on huge amount of data, which can be easily generated for a large variety of scenarios. While developments in both algorithms and computational capabilities have greatly improved the performance of DL and RL, the bottleneck in many cases remains the quality of the training datasets.

There are some basic requirements for a serial decision problem to fit into an RL framework, all of which are met when considering indoor mapping. First, is the ability to collect or simulate a lot of typical data representing the problem. One of the reasons for the great success of RL in gaming is the ability to simulate huge amounts of totally typical data. Fortunately, there are available datasets of building floor plans and, more importantly, such datasets can be generated. The second condition is that the problem must be Markovian, i.e. the outcome of an action depends only on the current state. Any serial decision problems that complies with this rule is called a Markov Decision Process (MDP). In some cases it is pertinent to encapsulate more than one state in order to obtain a good policy. One way to address this, while preserving the MDP formulation, is with an appropriate definition of the state. For instance, in our case, where the observations constitute the main element in representing the state, the current observation is insufficient. Rather, the state should include the accumulation of all past and current observations. Finally, even though the number of states in which the agent may be present is very large, defining a relatively small number of possible actions better facilitates the training process.

\section{RELATED WORK}
Until recently a leading method for indoor mapping was the frontier-based approach \cite{ymauchi}\cite{shrestha2019learned}, in which the agent searches forward towards the frontier between the explored regions and the unexplored ones. The main task of the algorithm designer was to choose the appropriate target location on the frontier. This choice is done greedily and is controlled by tuning the utility-based function, that balances between the expected area to be exposed at the selected point, and the distance to this point.
Following the remarkable results deep reinforcement learning (DRL) has shown when dealing with complicated problems, such as video games and board games \cite{mnih2015humanlevel} \cite{silver2016mastering} \cite{silver2017mastering} \cite{badia2020agent57}, sample-based approaches, mainly DRL-based methods, have become a popular choice when tackling the indoor mapping problem \cite{barratt2017active}\cite{chen2019self}. 
In learning navigation policies through DRL various strategies are explored by actively interacting with the environment, bringing about further breakthroughs in autonomous navigation.

In \cite{barratt2017active}\cite{chen2019self}\cite{botteghi2020reinforcement} a path planner is trained with DRL using a deep Q-network (DQN) or the Advantage Actor-Critic (A2C) algorithm, with a reward function based on the exposed area or on mission completeness. In these publications, the results are slightly weaker than frontier-based methods in terms of number of steps (or time). However, in \cite{chen2019self} the authors state that the decision-making itself using DRL-based algorithms may be faster when scaling up to larger domains. This is due to the fact that neural nets provide a constant calculation time compared to frontier-based methods that use heavy graph search. Hence, the main gain of using DRL stated in \cite{chen2019self} is related to computational issues and is not a result of statistical considerations. In particular, the DRL results in \cite{chen2019self} are similar to corresponding classic results, in terms of number of required actions while we seek to show that DRL outperforms Frontier-based methods.
In \cite{niroui2019deep} DRL is used to learn a point selection strategy as part of a frontier-based exploration algorithm. The policy network was trained using the Asynchronous Advantage Actor-Critic (A3C) algorithm. It was shown that this setup is superior to several variations of the cost-utility frontier-based method.Anyway, the suggested method their is a combination of frontier-based and reinforcement learning and not pure DRL one.

In \cite{ramezani2018end} the authors presented an end-to-end obstacle avoidance and navigation system based on DRL. They show that using a continuous action space as well as defining the state to be the last three observations (and not only the last one) improve agent’s performance. However, since their work focuses on obstacle avoidance and not on overall mapping of a delimited area, it cannot be compared to frontier-based algorithm. Similar works are \cite{barratt2017active} which focuses on outdoor mapping (whose statistics characteristics are different) and \cite{chen2019self} which focuses on navigation. Hence both cannot be compared to the frontier-based methods.

Another way to incorporate the known statistics of the environment is by using ML methods to obtain a prediction of the architecture in the unseen regions. Such a method is presented in \cite{shrestha2019learned}, where a variational autoencoder (VAE) is used as a map completion tool. In this work the required mapped area is segmented and each part is mapped sequentially, while the map completion network is used to select a frontier target which should be actually mapped. In previous work \cite{hayoun2020integrating} we use a map completion deep network to enhance mapping, while the mapping itself is done relying on the map completion module.

In this paper we integrate both the map completion module and the RL module to enhance the mapping process.


\section{PRELIMINARIES}
An MDP is a 4-tuple $(\mathcal{S}, \mathcal{A},\mathcal{P}_{a},\mathcal{R}_a)$. $\mathcal{S}$ is the state space and $\mathcal{A}$ is the action space. $\mathcal{P}_{a}$ represents the stochastic dynamics of the process (i.e. $\mathcal{P}_{a}(s,s')$ is the conditional probability of transition from state $s$ to state $s'$ after taking action $a$). $\mathcal{R}_a\left(s,s'\right)$ is the temporal reward for the transition from state $s$ to state $s'$ following action $a$.
A policy ${\pi}:\mathcal{S}\rightarrow\mathcal{A}$ is a function that maps a state to a subsequent action. The goal is to find a policy that maximizes the expected sum of decaying rewards
\begin{equation}
\label{eq:rl}
E\left[\sum_{t=0}^{\infty}\gamma^t\mathcal{R}_{\pi(s_t)}\left(s_t,\mathcal{P}_{\pi(s_t)}(s_t)\right)\right],
\end{equation}
where $\gamma\in\left[0,1\right]$ is the discount factor that exponentially decreases distant future rewards that are less guaranteed due to the uncertainty of the process. 

MDPs may be possible to solve analytically if the state space is small and the dynamics of the process are known. However, in many practical problems, such as ours, both conditions are not met.

\section{PROBLEM FORMULATION}
We consider the problem of mapping an unknown building, the boundary of which is known a priori, by an autonomous agent equipped with $360^o$ peripheral limited range sensors. The problem is simplified by neglecting any measurement noise and by assuming that the localization problem is solved. The structure to be mapped is regarded as a $2D$ occupancy grid, in which each cell may be either free space or an obstacle. The agent’s objective is to minimize the time to expose a desired portion of the map. At each time step the agent can move to any of its eight neighbors in the grid.

In accordance with the MDP formulation discussed in the previous section, we define the state $s$ as a discrete $2D$ map of dimensions $h \times w$ in which each cell $(x,y)$ can have one of four distinct values
\begin{equation}
\label{eq:thresholds}
s(x,y)=\begin{cases}
c_{\text{free}}, & \text{if $(x,y)$ is observed free space}\\
c_{\text{obstacle}} & \text{if $(x,y)$ is observed obstacle}\\
c_{\text{unknown}} & \text{if $(x,y)$ is unobserved}\\
c_{\text{agent}} & \text{the agent's location}.
\end{cases}
\end{equation}
In the following sections these four values are considered hyperparameters that should be chosen carefully (e.g. $c_{\text{agent}}$ should be dominant enough so that learning-based methods are able to easily detect its uniqueness).

The set of possible actions is given by
\begin{equation}
\label{eq:actions_space}
\mathcal{A}\in\{\mathcal{\uparrow,\nearrow,\rightarrow,\searrow,\downarrow,\swarrow, \leftarrow,\nwarrow}\}.
\end{equation}






 \section{DRL ALGORITHM DESIGN}
Problems such as indoor mapping, in which the state space is extremely large and the dynamics of the process are unknown, are impossible to solve analytically. Fortunately, as long as the state space can be represented in a compact way (e.g. matrix), a policy that optimizes the MDP objective function \eqref{eq:rl} can be derived using Q-learning or policy gradient methods \cite{sutton2018reinforcement}. Essentially, both approaches try to learn a parametric function that describes the "quality" of an action in a specific state. Up until the last decade, most of the ML community tried using convex representations of this function that are optimized easily, though less expressive. In recent years it has become commonplace to use neural networks -- a much more expressive, albeit non-convex, hypotheses class.

Several design choices need to be made when implementing an RL algorithm. First, the state and reward function must be delineated. Second, a specific algorithm and parametric architecture should be selected. Finally, the algorithm's hyper-parameters need to be tuned.

 
\subsection{State and Memory}
We used a grayscale (single channel) image to represent the state $s$. We found that the best values to optimize learning speed maintain $c_{\text{free}}<c_{\text{unknown}}<c_{\text{obstacle}}\ll c_{\text{agent}}$ (we used $c_{\text{free}}$=$0$, $c_{\text{unknown}}$=$15$, $c_{\text{wall}}$=$30$, and $c_{\text{agent}}$=$255$). We also discovered that fixing the agent to the center of the image, as in \cite{chen2019self}, significantly improves the results. Regarding the use of memory, we did not observe additional value by storing more than the last state. This is understandable, since state derivatives are inconsequential to the mapping process.
 
\subsection{Reward Function} 
Two conceptually different reward functions can be used: one that yields all the reward at the end of an episode (sparse) or one that provides a series of temporal rewards. The authors of \cite{botteghi2020reinforcement} compared these two options and concluded that the second is preferable. This is not surprising, since rewarding momentary partial exposures encourages the agent to expose more cells, hence speeding up the learning process.

\subsubsection{Penalties}
An important ingredient in the reward function is the compensation for completing the mapping task in a swift and safe manner. The temporal style of this component comes in the form of deduction. In our case the agent is penalized by a constant for each step taken, motivating it to complete the mapping as quickly as possible. In order to effectively train it to navigate safely, the agent is penalized by a relatively large value $\ell>1$ for actions that will lead it to collide with an obstacle (in training such actions are not executed to enable an episode's persistence).

\subsubsection{Rewards}
Following each action the agent receives an immediate reward proportional to the total size of any newly exposed areas. Although this enabled the agent to learn quite easily how to map most of the building, it still struggled to expose the last portions that might be far away from its location. To help the agent in learning to expose those challenging parts of the environment a non-stationary reward was added, which increases with the size of the area mapped so far. Choosing a convex function for this purpose ensured that the agent's training would be effective even as the amount of unexposed cells decreased.

The final reward function was chosen to be
\begin{equation}
 \label{eq:non_stationary_reward}
 \mathcal{R}_{a_t}(s_{t},s_{t+1})=-1 + \begin{cases}
 -\ell,~\text{if $a_t$ leads to collision}\\
 d \cdot n_{t+1} \cdot E^4(s_{t+1}),~~~~\text{else}.
 \end{cases}
\end{equation}
Here $d$ is a normalizing coefficient, $n_{t+1}$ is the number of cells that were exposed following action $a_t$, and $E(s_t)\in{\left[0,1\right]}$ is the ratio between the exposed area in $s_{t+1}$ and the total area of the building.

\subsection{Algorithm} 
Several algorithms of both Q-learning and policy gradient methods were examined, namely: DQN, A2C, and Proximal Policy Optimization (PPO). Of those we found that PPO yielded the best results, both in terms of training duration and in terms of the quality of the results, hence it was our RL algorithm of choice.

\subsection{Architecture} 
The CNN introduced in \cite{mnih2015humanlevel} was used in the learning of the policy. The output layer of the network includes nine scalars: eight for the policy distribution (actor) and an additional ninth which estimates the value function (critic). In this architecture, the networks of the value function and the policy distribution are the same. Several other network architectures were examined (such as MLP and some variations of the CNN network), but none yielded better results. 

\section{DL-BASED MAP PREDICTION}

\begin{figure}[b]
  \includegraphics[width=\linewidth]{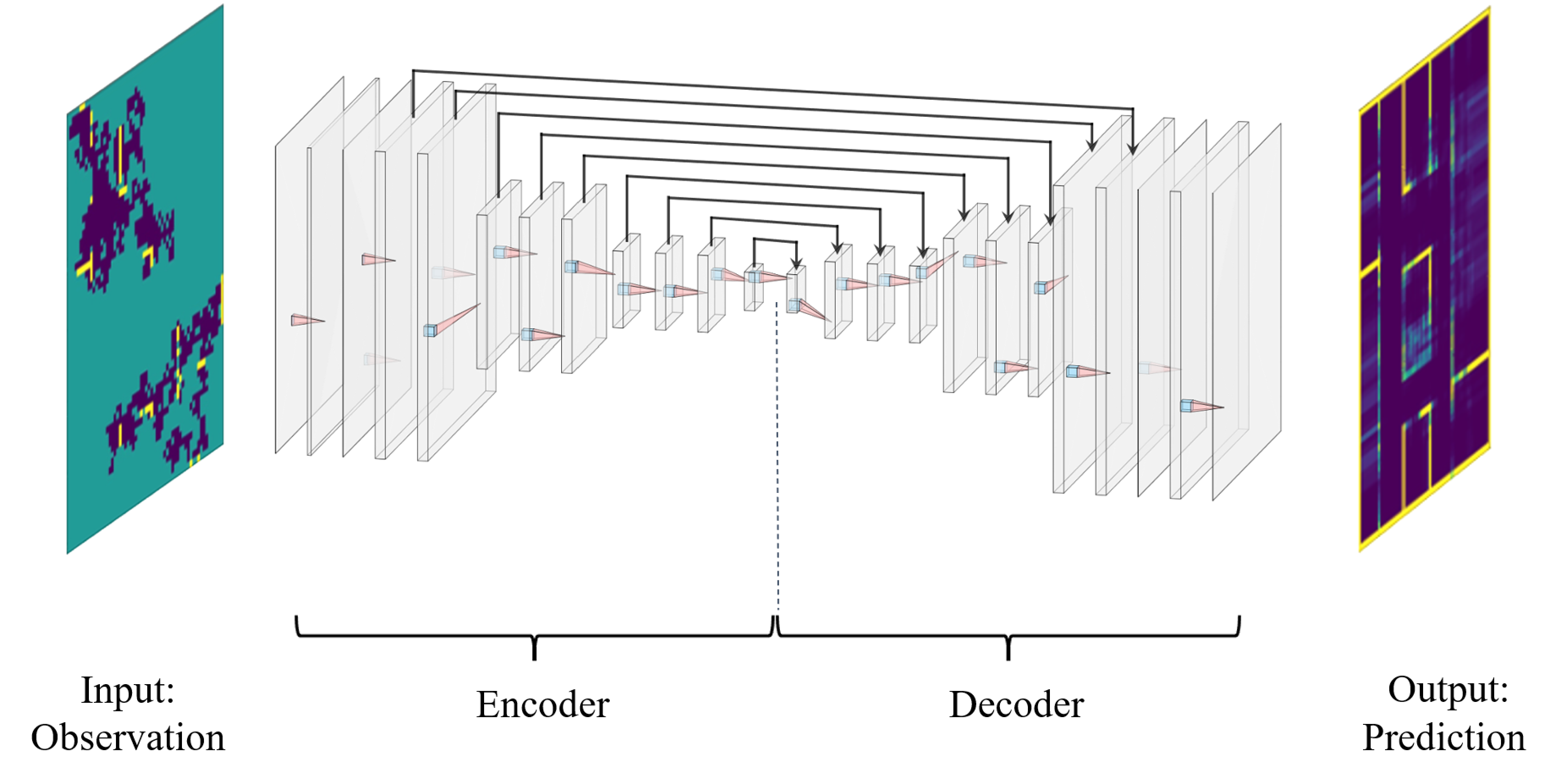}
  \caption{Autoencoder-based map predictor. The neural network is symmetric with eleven convolution encoder layers and eleven deconvolution decoder layers. Each encoder layer is additionally connected to its counterpart in the decoder. Input - a partial observation of a building. Output - the probability for each cell of being a wall.}
  \label{fig:nn_architecture}
\end{figure}

\begin{figure}[t]
\vspace{10pt}
  \includegraphics[width=\linewidth]{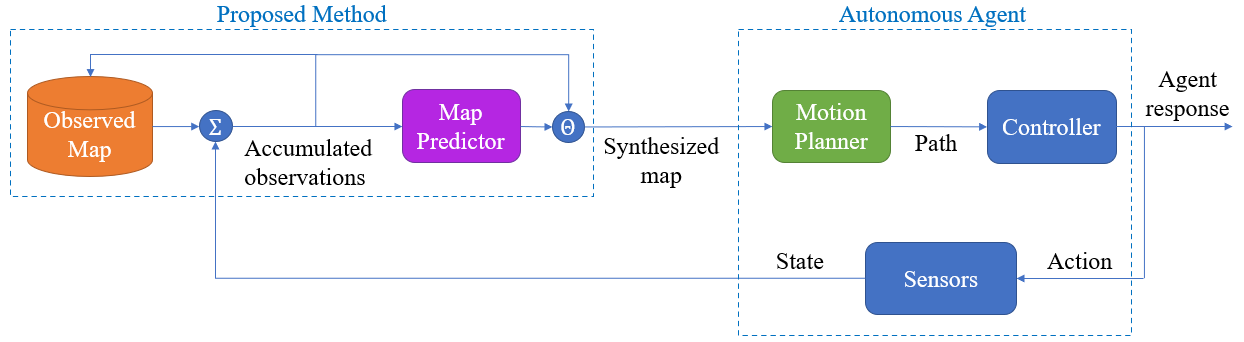}
  \caption{The proposed cascaded control scheme of the our agent. On the left hand side is our contribution including $\Sigma$ - the observations accumulation operator, and $\Theta$ - the observations and prediction synthesis operator. On the right hand side is a common control scheme of an autonomous agent.
  }
  \label{fig:block_diagram}
\vspace{-10pt}
\end{figure}

Another mode by which the underlying statistics of the environment can be reflected is the inclusion of an DL-based map predictor, developed by the authors in previous work \cite{hayoun2020integrating}. The predictor itself is a composition of two functions $f_{\text{threshold}} \circ f_{\text{prediction}}$, where $f_{\text{prediction}}$ represents an obstacle-predicting network and $f_{\text{threshold}}$ is a thresholding function. $f_{\text{prediction}}$ is given by a ResNet-styled convolutional autoencoder (see Figure \ref{fig:nn_architecture}). Auotencoders are effectively utilized for tasks such as anomaly detection \cite{an2015variational} and image restoration \cite{mao2016image}. In our application the network is trained to output a probabilistic estimation of the complete map, given partial observations. It is essentially a function from the observation space to the space of obstruction probability maps
\begin{equation}
    f_{\text{prediction}}:\mathcal{O} \rightarrow \left[0,1\right]^{h\times w},
\end{equation}
where $\mathcal{O}=\{c_{\text{free}},c_{\text{obstacle}},c_{\text{unknown}}\}^{h\times w}$ is the observation space, and where the output value $0$ indicates certain free space and $1$ indicates certain obstacles.
The thresholding function $f_{\text{threshold}}$ maps the probabilistic output from $f_{\text{prediction}}$, denoted by $p_m \in \left[0,1\right]^{h\times w}$, back to the observation space. It does so with the use of two confidence levels, $\delta_{\text{free}}$ and $\delta_{\text{obstacle}}$, in the following manner:
\begin{equation}
\label{eq:thresholds}
f_{\text{threshold}}\left(p_m(x,y)\right)=\begin{cases}
c_{\text{free}}, & p_{m}(x,y) \leq \dfrac{1-\delta_{\text{free}}}{2}\\
\vspace{-8pt}\\
c_{\text{obstacle}}, & p_{m}(x,y) \geq \dfrac{1+\delta_{\text{obstacle}}}{2}\\
c_{\text{unknown}}, & \text{otherwise}.
\end{cases}
\end{equation}
The thresholds actually determine the trade-off between the number of false positives and negatives and the map construction rate. The chosen set of values was obtained through trial and error until a good balance between the thresholded prediction's $F_1$-score and the mapping duration was found.
Thus we obtain the map predictor 
\begin{equation}
    f_{\text{predictor}} = f_{\text{threshold}} \circ f_{\text{prediction}}: \mathcal{O} \rightarrow \mathcal{O}
\end{equation}
that can be incorporated into the RL loop, as illustrated in Figure \ref{fig:block_diagram}. The final prediction-based map is synthesized by overlaying the observed sections of the map.

In effect, the predictor acts as a learned estimate of the state transition function (dynamics) of the MDP to be solved. As such, it also affects the reward function by providing foresight with respect to future exposure. Therefore, one might consider the combined setup in which the DRL-based motion planner is trained with the predictor as model-based. Furthermore, assuming the prediction chosen thresholds ensure a reasonable $F_1$-score, the augmented observations map can also serve as the outputted constructed map. This provides an additional means to shorten the mapping duration, by expanding the uncovered areas at any given time.

\section{Simulations}

\subsection{Simulation Testbed}
A simplified grid world simulation was set up in Python, in which the mapping agent, situated in a certain cell, is free to move to any of its eight neighbouring cells, provided they are vacant. The agent is equipped with $16$ fixed on-board range sensors arranged in equal angular intervals of $22.5^{\circ}$ with an effective range of $20$ cells.

\subsection{Training Sets}

\begin{figure}[b]
  \includegraphics[width=\linewidth]{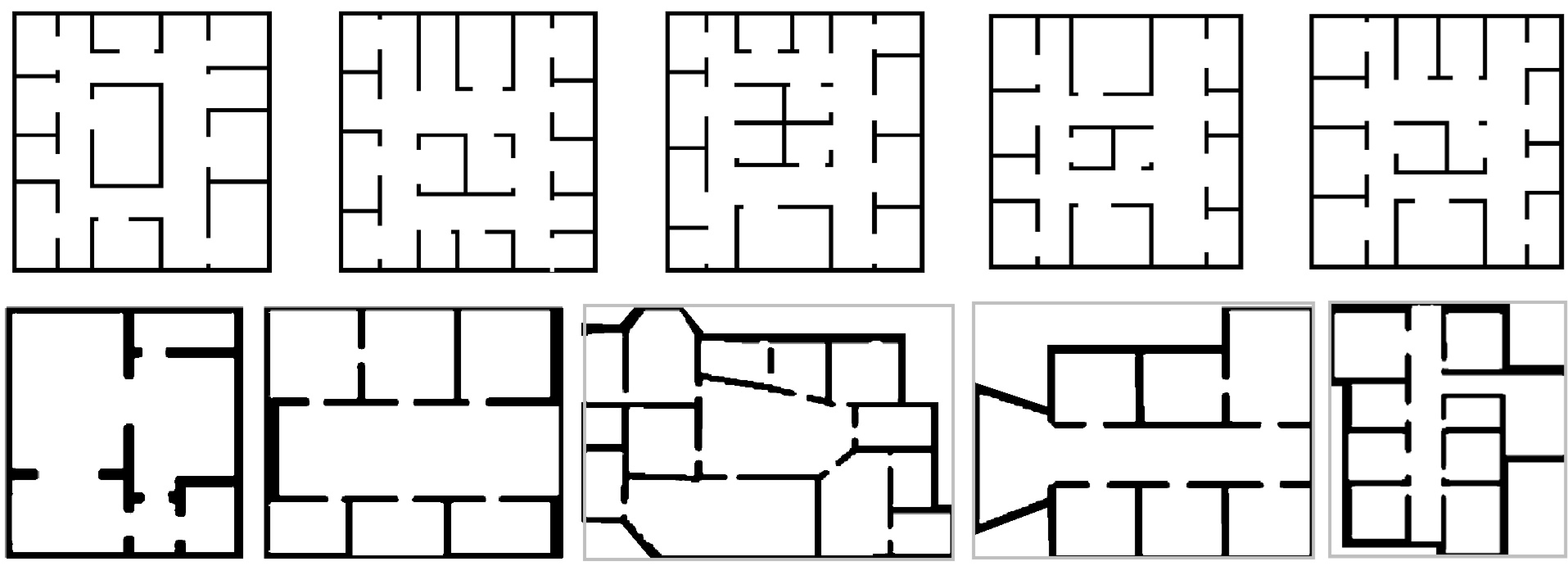}
  \caption{Illustrative floorplan examples. The top row includes examples from $\mathcal{D}_1$ and the bottom row includes examples from $\mathcal{D}_2$}
  \label{fig:dbs}
\end{figure}

\begin{table}[b]
\caption{Datasets characteristics}
\label{table:datasets}
\begin{center}
\begin{tabular}{clll}
\cline{3-4}
\multicolumn{1}{l}{} & \multicolumn{1}{l|}{} & \multicolumn{2}{c|}{\multirow{2}{*}{\textit{\textbf{\size{10}{Datasets}}}}} \\
\multicolumn{1}{l}{} & \multicolumn{1}{l|}{} & \multicolumn{1}{l}{} & 
\multicolumn{1}{l|}{} \\
\hhline{~~|==|} 
\multicolumn{1}{l}{} & \multicolumn{1}{l|}{} & \multicolumn{1}{c|}{\multirow{1}{*}{$\mathcal{D}_1$}} & 
\multicolumn{1}{c|}{\multirow{1}{*}{$\mathcal{D}_2$}} \\
\hline 
\hhline{|~||~|--} 
\hhline{|~||~|--}
\multicolumn{1}{|c||}{\multirow{5}{*}{\rotatebox[origin=c]{90}{\parbox[c]{0.9cm}{\centering \textit{\textbf{\size{10}{\begin{tabular}[c]{@{}c@{}} Map \\ Features \end{tabular}}}}}}}} & \multicolumn{1}{l|}{\multirow{1.2}{*}{Contour}} & \multicolumn{1}{?l|}{\multirow{1.2}{*}{Convex (rectangle)}} & 
\multicolumn{1}{l?}{\multirow{1}{*}{Convex\slash Concave}} \\ 
\cline{2-4} 
\multicolumn{1}{|c||}{\multirow{1.2}{*}{}} & \multicolumn{1}{l|}{\multirow{1.2}{*}{Size}} & \multicolumn{1}{?l|}{\multirow{1.2}{*}{3,720 (0) cells}} &
\multicolumn{1}{l?}{\multirow{1.2}{*}{2,140 (1,100) cells}} \\
\cline{2-4} 
\multicolumn{1}{|c||}{} &  \multicolumn{1}{l|}{\multirow{1.2}{*}{\%Walls}} & \multicolumn{1}{?l|}{\multirow{1.2}{*}{7.3\% (0.4\%)}} &
\multicolumn{1}{l?}{\multirow{1.2}{*}{3.9\% (2.0\%)}} \\
\cline{2-4} 
\multicolumn{1}{|c||}{} & \multicolumn{1}{l|}{\multirow{1.2}{*}{Topology}} & \multicolumn{1}{?l|}{\multirow{1.2}{*}{Similar}} & 
\multicolumn{1}{l?}{\multirow{1.2}{*}{Diverse}} \\
\hhline{|~||~|--} 
\hhline{|~||~|--} 
\hline
\end{tabular}
\end{center}
\end{table}

The map prediction and motion planning networks were trained on two distinct datasets, dubbed $\mathcal{D}_1$ and $\mathcal{D}_2$. Dataset $\mathcal{D}_1$ holds $50,000$ independently generated maps, and $\mathcal{D}_2$ contains $15,894$ maps from the HouseExpo dataset in \cite{li2020houseexpo}. Several examples from each dataset are displayed in Figure \ref{fig:dbs}. Table \ref{table:datasets} shows a qualitative and statistical comparison of the geometrical features that characterize each dataset, highlighting the diversity of $\mathcal{D}_2$.
This fact is later shown to influence the mapping success rate, as summarised in Table \ref{table:final_res}. 

After training the predictor the thresholds were tuned. We chose $\delta_{\text{free}}=0.93, ~\delta_{\text{obstacle}}=0.95$ for $\mathcal{D}_1$ and $\delta_{\text{free}}=0.9, ~\delta_{\text{obstacle}}=0.99$ for $\mathcal{D}_2$, which yielded a minimal $F_1$-score of $0.92$.

Separate DRL-based motion planners were trained for different values of the required exposure percentage. In the following we will present results for $85\%$ and $98\%$.

\subsection{Simulation Results \& Analysis}

\begin{table}[b]
\caption{Relative Mapping Duration Reductions}
\label{table:final_res}
\begin{center}
\begin{tabular}{cccccc}
\cline{4-6}
& & \multicolumn{1}{c|}{} & \multicolumn{3}{c|}{\multirow{2}{*}{\textit{\textbf{\size{10}{Motion Planner}}}}} \\ 
& & \multicolumn{1}{c|}{} & \multicolumn{3}{c|}{} \\ 
\hhline{~~~|===|} 
& & \multicolumn{1}{c|}{} & \multicolumn{1}{c|}{PPO} & \multicolumn{1}{c|}{\begin{tabular}[c]{@{}c@{}}Frontier- \\ based \\ with \\ prediction \end{tabular}} & \multicolumn{1}{c|}{\begin{tabular}[c]{@{}c@{}}PPO \\ with \\ prediction \end{tabular}} \\ 
\cline{1-6}
\hhline{|~||~|~|---} 
\hhline{|~||~|~|---} 
\multicolumn{1}{|c||}{\multirow{10}{*}{\rotatebox[origin=c]{90}{{\textit{\textbf{\size{10}{Desired Exposure}}}}}}} & \multicolumn{1}{c|}{\multirow{4}{*}{85\%}} & \multicolumn{1}{c|}{$\mathcal{D}_1$} & \multicolumn{1}{?c|}{\begin{tabular}[c]{@{}c@{}}32.10\% \\ (11.63\%) \\ {[100\%]}\end{tabular}} & \multicolumn{1}{c|}{\begin{tabular}[c]{@{}c@{}}75.31\% \\ (5.07\%) \\ {[100\%]}\end{tabular}} & \multicolumn{1}{c?}{\begin{tabular}[c]{@{}c@{}}\textbf{77.59\%} \\ \textbf{(4.05\%)} \\ {[100\%]}\end{tabular}} \\ 
\cline{3-6}
\multicolumn{1}{|c||}{} & \multicolumn{1}{c|}{} & 
\multicolumn{1}{c|}{$\mathcal{D}_2$} & \multicolumn{1}{?c|}{\begin{tabular}[c]{@{}c@{}}12.07\% \\ (20.22\%) \\ {[96\%]}\end{tabular}}  & \multicolumn{1}{c|}{\begin{tabular}[c]{@{}c@{}}57.75\% \\ (18.09\%) \\ {[100\%]}\end{tabular}} & \multicolumn{1}{c?}{\begin{tabular}[c]{@{}c@{}}\textbf{61.90\%} \\ \textbf{(16.77\%)} \\ {[99\%]}\end{tabular}} \\ 
\cline{2-6}
\multicolumn{1}{|c||}{} & \multicolumn{1}{c|}{\multirow{4}{*}{98\%}} & 
\multicolumn{1}{c|}{$\mathcal{D}_1$} & \multicolumn{1}{?c|}{\begin{tabular}[c]{@{}c@{}}21.91\% \\ (9.88\%) \\ {[97\%]}\end{tabular}} & \multicolumn{1}{c|}{\begin{tabular}[c]{@{}c@{}}71.80\% \\ (5.12\%) \\ {[100\%]}\end{tabular}} & \multicolumn{1}{c?}{\begin{tabular}[c]{@{}c@{}}\textbf{75.18\%} \\ \textbf{(4.21\%)} \\ {[97\%]}\end{tabular}} \\ 
\cline{3-6}
\multicolumn{1}{|c||}{} & \multicolumn{1}{c|}{} & \multicolumn{1}{c|}{$\mathcal{D}_2$} & \multicolumn{1}{?c|}{\begin{tabular}[c]{@{}c@{}}5.11\% \\ (25.16\%) \\ {[87\%]}\end{tabular}}  & \multicolumn{1}{c|}{\begin{tabular}[c]{@{}c@{}}57.28\% \\ (16.69\%) \\ {[99\%]}\end{tabular}} & \multicolumn{1}{c?}{\begin{tabular}[c]{@{}c@{}}\textbf{63.78\%} \\ \textbf{(14.24\%)} \\ {[91\%]}\end{tabular}} \\ 
\hhline{|~||~|~|---} 
\hhline{|~||~|~|---} 
\cline{1-6}
\end{tabular}
\end{center}
\end{table}

Three proposed motion planning algorithms were evaluated and compared to the baseline cost-utility variation of the frontier-based planning presented in \cite{hayoun2020integrating}: frontier-prediction-based, model-free PPO, and model-based PPO.

In order to assess the different algorithms we ran them on 100 maps from each dataset with different desired exposure amounts. These maps were excluded from the training set of the PPO algorithms. The success criteria was reaching the exposure requirement in under $400$ steps. A summary of the performance of each algorithm over the simulation runs in which all algorithms succeeded is shown in Table \ref{table:final_res}. The results are given in terms of the achieved reduction in mapping time relative to the frontier-based algorithm. The upper value in each cell is the time-saving percentage, the values in round brackets are the standard deviations, and the values in the squared brackets are the success rates over all runs. In the sequel we discuss the failed mapping instances.

\begin{figure}[ht]
\vspace{10pt}
     \centering
     \begin{subfigure}[b]{0.3\linewidth}
         \centering
         \captionsetup{width=\linewidth}
         \includegraphics[width=\linewidth]{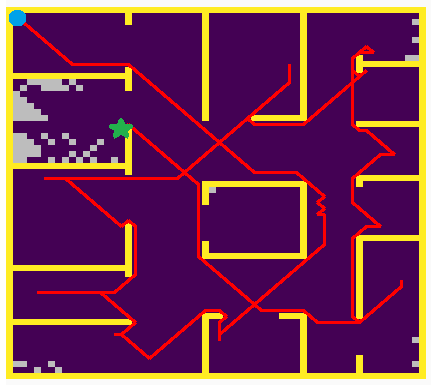}
         \caption{Frontier-based path example in $\mathcal{D}_1$ ($323$ steps)}
         \label{fig:classic_db1}
     \end{subfigure}
     \hfill
     \begin{subfigure}[b]{0.3\linewidth}
         \centering
         \captionsetup{width=\linewidth}
         \includegraphics[width=\linewidth]{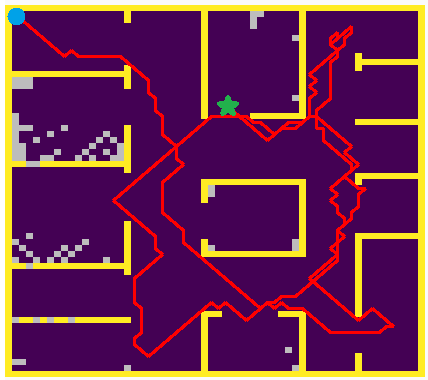}
         \caption{PPO path example in $\mathcal{D}_1$, without the predictor ($267$ steps)}
         \label{fig:ppo_wihtout_db1}
     \end{subfigure}
     \hfill
     \begin{subfigure}[b]{0.3\linewidth}
         \centering
         \captionsetup{width=\linewidth}
         \includegraphics[width=\linewidth]{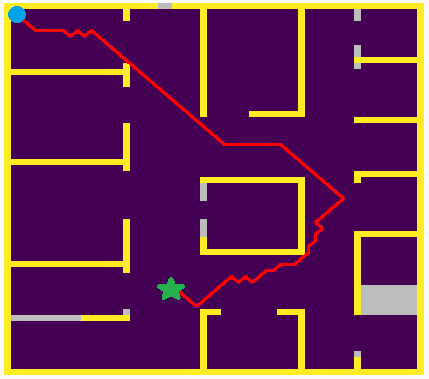}
         \caption{PPO path example in $\mathcal{D}_1$, with the predictor   ($75$ steps)}
         \label{fig:ppo_with_db1}
     \end{subfigure}
     
     \bigskip
     
     \begin{subfigure}[b]{0.3\linewidth}
         \centering
         \captionsetup{width=\linewidth}
         \includegraphics[width=\linewidth]{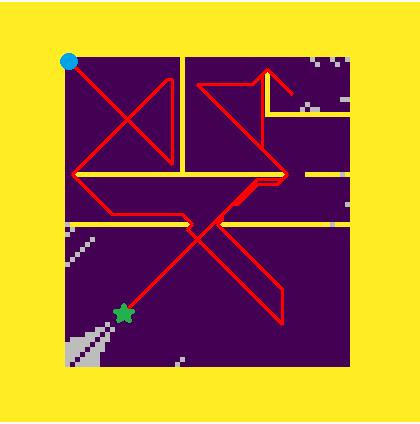}
         \caption{Frontier-based path example in $\mathcal{D}_2$ ($233$ steps)}
         \label{fig:classic_db2}
     \end{subfigure}
     \hfill
     \begin{subfigure}[b]{0.3\linewidth}
         \centering
         \captionsetup{width=\linewidth}
         \includegraphics[width=\linewidth]{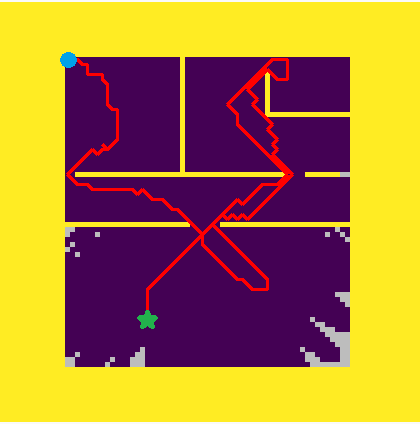}
         \caption{PPO path example in $\mathcal{D}_2$, without the predictor ($195$ steps)}
         \label{fig:ppo_without_db2}
     \end{subfigure}
     \hfill
     \begin{subfigure}[b]{0.3\linewidth}
         \centering
         \captionsetup{width=\linewidth}
         \includegraphics[width=\linewidth]{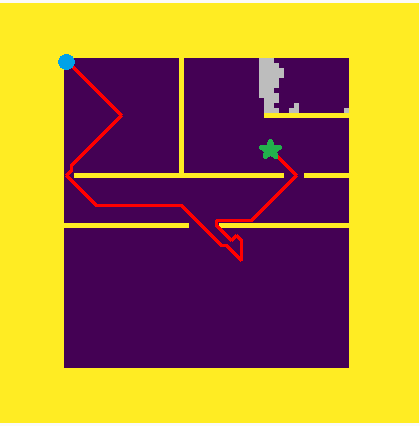}
         \caption{PPO path example in $\mathcal{D}_2$, with the predictor    ($89$ steps)}
         \label{fig:ppo_with_db2}
     \end{subfigure}
     \captionsetup{width=\linewidth}
     \caption{Paths generated by the frontier-based method and PPO with and without predictor, for representative examples from our two datasets. The required exposure rate in all cases is $98\%$. The traced paths, in red, start from the blue circle (top left corner) and end at the green star.}
     \label{fig:paths}
\end{figure}

Our main conclusions from this analysis are: \\
\textbf{Predictor contribution:} The predictor improves dramatically the performance of the mapping procedure. For both datasets and in both required coverage the predictor reduces the required mapping time by $60\%$ to $75\%$. It shows the strength of the predictor including for the $\mathcal{D}_2$'s diversed buildings case. In $\mathcal{D}_1$ and $98\%$ required coverage while using PPO and without predictor, normalized (number of steps per $1,000$ pixels in grid) mean required mapping steps is $\sim$66 and it reduces to $\sim$21 while adding predictor. In $\mathcal{D}_2$ adding predictor reduces the normalized mean required mapping duration from $\sim63$ to $\sim28$ steps. \\
\textbf{DRL planner contribution:} Without predictor, mapping using PPO is equal and even better comparing to the frontier-based method, in $10$ to $50$ percents (for completed episodes), while the difference is higher for the well ordered dataset $\mathcal{D}_1$ and lower for $\mathcal{D}_2$. These results seams to improve the reported results in~\cite{barratt2017active}\cite{chen2019self}\cite{botteghi2020reinforcement} where RL was nearly optimal. Yet, the predictor's contribution is the primary and DRL's contribution is the secondary. \\
\textbf{Datasets uniformity:} In the ordered $\mathcal{D}_1$ the advantage of PPO over the frontier-based is much more significant, see Table \ref{table:final_res}. This is not surprising: $\mathcal{D}_1$ is subjected to much more strict statistics, so learning from examples is much more officiant. However, we assume that if we had bigger dataset to train $\mathcal{D}_2$, the RL model could utilize it to understand the statistics of the buildings better for achieving better performance. RL algorithms try to optimize the sum of the overall decaying rewards while the frontier-based algorithm looks greedily for the optimal step. This difference leads to changes between the path created by the various algorithms, as depicted in Figure \ref{fig:paths}. The frontier-based algorithm goes in straight lines along walls and toward corners entering side rooms, while the path caused by PPO zigzagging mainly in the middle of corridors. \\
\textbf{Failures:} We experienced some failure episodes, caused by two reasons: 
First, the predictor may suggest a room to be inaccessible with a false positive error of identifying a space as a wall, while the agent is inside that "closed" room. In such a case, both frontier-based and RL algorithms failed.
Second, RL algorithm may fail and go in circles, till the time allocated to the episode is finished, and we relate to this phenomena in Section \ref{Conclusions}. In our evaluation, the fraction of failures episodes was relatively low: not more than $4$ percents for all configurations and datasets, excluding the case of $\mathcal{D}_2$ with required coverage of $98\%$, in which the failure rate was $13\%$ without predictor and $9\%$ with predictor. Note that the failures stated in Table \ref{table:final_res} are not really total failures: even for the unsuccessfully episodes, final coverage was about $90\%$.

\begin{figure}
\vspace{10pt}
     \centering
     \begin{subfigure}[b]{\linewidth}
         \centering
         \includegraphics[width=0.75\linewidth]{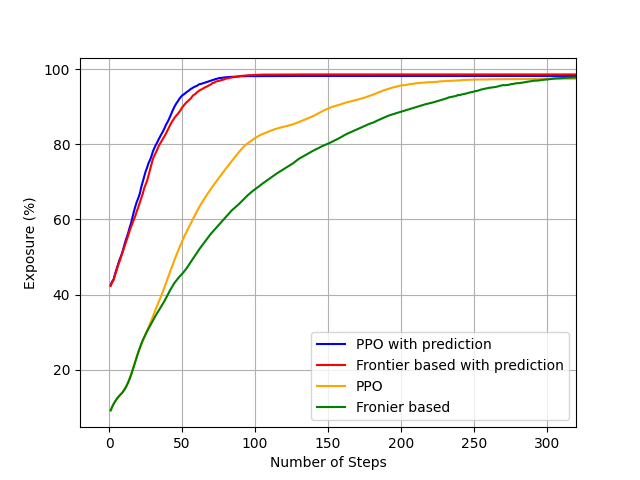}
         \caption{Exposure propagation in $\mathcal{D}_1$}
         \label{fig:db1_stat}
     \end{subfigure}

     \begin{subfigure}[b]{\linewidth}
         \centering
         \includegraphics[width=0.75\linewidth]{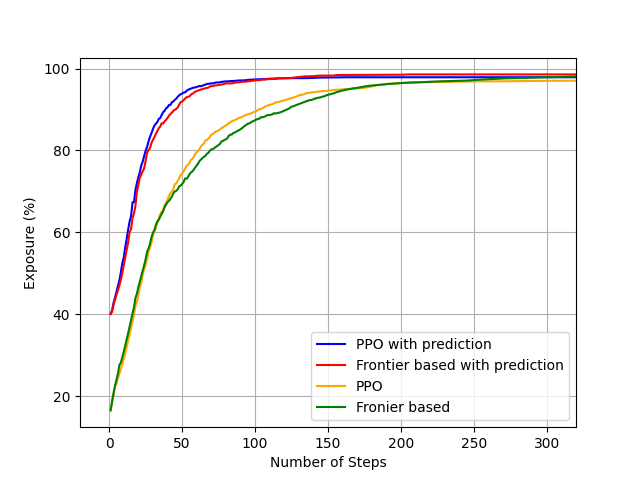}
         \caption{Exposure propagation in $\mathcal{D}_2$}
         \label{fig:db2_stat}
     \end{subfigure}
     \caption {Average exposure propagation over each dataset for a target value of $98\%$. Note that before the exploration starts 
     the cells near the external walls are predicted as vacant, hence the exposure propagation starts with approximately 40\% coverage.}
     \label{fig:stat}
\end{figure}

In Figure \ref{fig:stat} we compare the exposure propagation of the mapping process for both  $\mathcal{D}_1$ and $\mathcal{D}_2$. As seen, PPO algorithm always yields better results than frontier-based, while the PPO combined with predictor returns the best results. The  improvements in total mapping time presented in Table \ref{table:final_res} is clearly seen: e.g. while in Figure \ref{fig:db1_stat} the frontier-based curve (in Green) converges at $\sim300$ steps, the PPO with predictor curve (Blue) converges at $\sim75$ steps.

\section{FUTURE WORK}

In future work we intend to implement the methods described in this paper in a lab setting. We also plan to extend the current problem to include multiple cooperative agents and to develop fitting DL and DRL-based strategies for such multi-agent systems.


\section{CONCLUSIONS} \label{Conclusions}
Two learning-based methods were proposed to address the indoor mapping problem in cases where statistical architectonic information about the environment is available. One is to train the motion planner through reinforcement, in an endeavor to incorporate the prior structural knowledge directly into the decision making process. The other is the inclusion of a map predictor capable of expanding the explored areas in the map. 
The PPO algorithm was chosen in order to train the motion planner, realized as a neural network. The map predictor -- a convolutional autoencoder -- was trained on partial observations in a supervised manner. Two motion planning architectures were examined: one including only the DRL-based planner and another incorporating the map predicting network as well. Several versions of both setups were produced after training on one of two distinct datatsets for different specified desired exposure percentages. Their performance was compared, through simulation, to two instances of a cost-utility frontier-based algorithm: one with the addition of the map predictor and one without.

Of the examined configurations the combination of the DRL-based motion planner and map predictor yielded the best results, in terms of the mapping duration, where the better part of the mapping time reduction was evidently achieved by the map predictor. However, with respect to the success rate, we found that in some instances the trained planner would reach a point of indecision. This would cause a jitter in the agent's path and, in extreme cases, failure to complete the mapping in the allotted time. In real-world applications, where the system cannot tolerate any misconduct, a frontier-based algorithm can also be integrated into the motion planning logic, along with the DRL-based planner. Thus, if a deadlock is reached the frontier-based planner can serve as a recovery mechanism (similar to~\cite{chen2019self}). In this way we can eliminate any misbehaviour by the trained motion planner and achieve a perfect success rate.

We have demonstrated how prior knowledge of the underlying statistics of a given problem can substantially improve its solution. 
In the case of indoor mapping we managed to achieve a significant reduction in the mapping duration by incorporating the environment's structural statistics into the motion planning process.
In light of the results presented in this paper, it is our view that utilization of any such available data should be a central component while designing autonomous agents.

\addtolength{\textheight}{-12cm}   



\bibliographystyle{IEEEtran} 
\bibliography{IEEEabrv,rl_based_mapping}

\end{document}